\documentclass[conference,transmag]{IEEEtran}
\usepackage{makecell}

\IEEEoverridecommandlockouts
\usepackage{cite}
\usepackage{amsmath,amssymb,amsfonts}
\usepackage{algorithmic}
\usepackage{algorithm}
\usepackage{graphicx}
\usepackage{textcomp}
\usepackage{xcolor}
\usepackage{url}
\def\BibTeX{{\rm B\kern-.05em{\sc i\kern-.025em b}\kern-.08em
    T\kern-.1667em\lower.7ex\hbox{E}\kern-.125emX}}

\begin{document}

\title{Graph-PiT: Enhancing Structural Coherence in Part-Based Image Synthesis via Graph Priors}

\author{
\IEEEauthorblockN{Junbin Zhang\IEEEauthorrefmark{1}, Meng Cao\IEEEauthorrefmark{2}, Feng Tan\IEEEauthorrefmark{1}, Yikai Lin\IEEEauthorrefmark{1}, Yuexian Zou\IEEEauthorrefmark{1,*}}
\IEEEauthorblockA{\IEEEauthorrefmark{1}\textit{Guangdong Provincial Key Laboratory of Ultra High Definition Immersive Media Technology}}
\IEEEauthorblockA{\textit{Shenzhen Graduate School, Peking University, China}}
\IEEEauthorblockA{\IEEEauthorrefmark{2}\textit{Department of Computer Vision, Mohamed bin Zayed University of Artificial Intelligence, Abu Dhabi, United Arab Emirates}}
\IEEEauthorblockA{junbinzhang@pku.edu.cn, mengcaopku@gmail.com, 2301212743@stu.pku.edu.cn, linyikai@stu.pku.edu.cn, zouyx@pku.edu.cn}
}

\maketitle

\begin{abstract}
\textit{Abstract—}Achieving fine-grained and structurally sound controllability is a cornerstone of advanced visual generation. Existing part-based frameworks treat user-provided parts as an unordered set and therefore ignore their intrinsic spatial and semantic relationships, which often results in compositions that lack structural integrity. To bridge this gap, we propose Graph-PiT, a framework that explicitly models the structural dependencies of visual components using a graph prior. Specifically, we represent visual parts as nodes and their spatial-semantic relationships as edges. At the heart of our method is a Hierarchical Graph Neural Network (HGNN) module that performs bidirectional message passing between coarse-grained part-level super-nodes and fine-grained IP+ token sub-nodes, refining part embeddings before they enter the generative pipeline. We also introduce a graph Laplacian smoothness loss and an edge-reconstruction loss so that adjacent parts acquire compatible, relation-aware embeddings. Quantitative experiments on controlled synthetic domains (character, product, indoor layout, and jigsaw), together with qualitative transfer to real web images, show that Graph-PiT improves structural coherence over vanilla PiT while remaining compatible with the original IP-Prior pipeline. Ablation experiments confirm that explicit relational reasoning is crucial for enforcing user-specified adjacency constraints. Our approach not only enhances the plausibility of generated concepts but also offers a scalable and interpretable mechanism for complex, multi-part image synthesis. The code is available at https://github.com/wolf-bailang/Graph-PiT.
\end{abstract}


\section{Introduction}
\label{sec:intro}
Controllable image synthesis has advanced dramatically with the advent of large-scale diffusion models and rich conditioning mechanisms~\cite{rombach2022high},\cite{radford2021learning},\cite{karras2019style},\cite{gu2023mix}. However, fine-grained compositional control remains challenging. In industrial design, 3D modeling, and creative AI, designers often assemble new concepts by arranging existing parts or sub-components (e.g., prototyping a robot from limbs, designing furniture layouts, or crafting novel characters). However, most diffusion-based generators assume text or global image prompts, ignoring explicit part relationships. For example, PiT (Piece it Together)\cite{richardson2025piece} encodes each part into an $IP^+$ token grid and composes the set through an IP-Prior diffusion model, but it still treats the parts as an unordered set, allowing components to be placed in implausible configurations. In contrast, many real-world domains require structured assemblies: mechanical parts must join at specific hinges, furniture components must align according to room geometry, and product elements, such as wheels and chassis, have fixed spatial adjacency. As noted by Chen et al.\cite{ruiz2023dreambooth} on 3D generation, assets typically consist of a single, fused representation without any useful structure. However, most applications and creative workflows require assets to be made of several meaningful parts.

To address this, we propose Graph-PiT, which endows part-based generative modeling with explicit structural priors. We represent each input part as a node in a graph and connect edges to encode spatial or semantic relationships (e.g., adjacency, attachment points, or user-specified constraints). A hierarchical Graph Neural Network (HGNN) propagates information along this graph so that inter-part constraints inform generation. The refined part embedding then conditions a latent diffusion model (e.g., a latent SDXL generator~\cite{podell2023sdxl}), ensuring that the output image not only matches each part’s appearance but also respects the desired layout. By design, Graph-PiT can take both image patches and a skeleton graph (or layout) as inputs and synthesize images that are globally coherent and physically plausible in the 2D part-layout sense considered here. In this paper, structural coherence refers to respecting adjacency and topology among visible parts rather than enforcing full 3D geometry, viewpoint, or occlusion consistency. In creative workflows, this allows, for instance, specifying that a chair’s legs attach at the seat and that robot arms connect to the torso, yielding more realistic and controllable results. As PartComposer~\cite{rangwani2025composing} demonstrates, artists need part-level control; current models often “ignore the attribute details” at the part level, and our graph-based approach closes this gap by encoding structural priors. Our contributions are threefold:

\begin{itemize}
    \item \textbf{Graph prior.} We introduce a two-tiered graph over input parts whose edges encode spatial/semantic constraints, derived either automatically during training or directly from user-specified layouts.
    \item \textbf{Hierarchical graph aggregator.} We design an HGNN that couples part-level super-nodes and token-level $IP^+$ sub-nodes token features via bidirectional top-down/bottom-up message passing and two structural regularizers (Laplacian smoothness and edge reconstruction), going beyond treating the graph simply as an extra token encoder.
    \item \textbf{Graph-conditioned diffusion prior compatible.} We condition the IP-Prior on the graph-aggregated tokens in a way that degenerates exactly to the original PiT model when graph tokens are removed, enabling a clean comparison, ablations, and drop-in replacement in existing PiT pipelines.
\end{itemize}

\section{Methodology}
In this section, we present Graph-PiT, a framework that elevates part-based image synthesis from an unordered collection of components to a structurally coherent generation process guided by relational priors. For related work, please refer to Supplementary Section V. For preliminary knowledge on $IP^+$ representation and GNN background, please refer to Supplementary Section VI.

\subsection{Problem Formulation}
\textbf{Part-Based Concept Generation.} Let $\mathcal{I} = \{I_1, I_2, \ldots, I_N\}$ be an unordered collection of $N$ part images (e.g., a chair's leg, backrest, and seat cushion). Our objective is to generate a complete concept $x$ (an embedding or rendered image) that plausibly integrates all provided parts while satisfying structural constraints encoded by a graph $\textbf{G}$. Formally, we seek a generative distribution
\begin{equation}
\textbf{x} \sim \textbf{p}_\theta(\textbf{x} | \{I_i\}_{i=1}^N, \textbf{G}) \label{eq:equation1}
\end{equation}
where $\mathbf{G} = (\textbf{V}, \textbf{E})$ is a part-relationship graph whose nodes $\textbf{V}$ correspond to the $N$ parts and whose edge set $\textbf{E}$ encodes spatial or semantic adjacency.

\begin{figure}[htbp]
\centerline{\includegraphics[width=.8\linewidth]{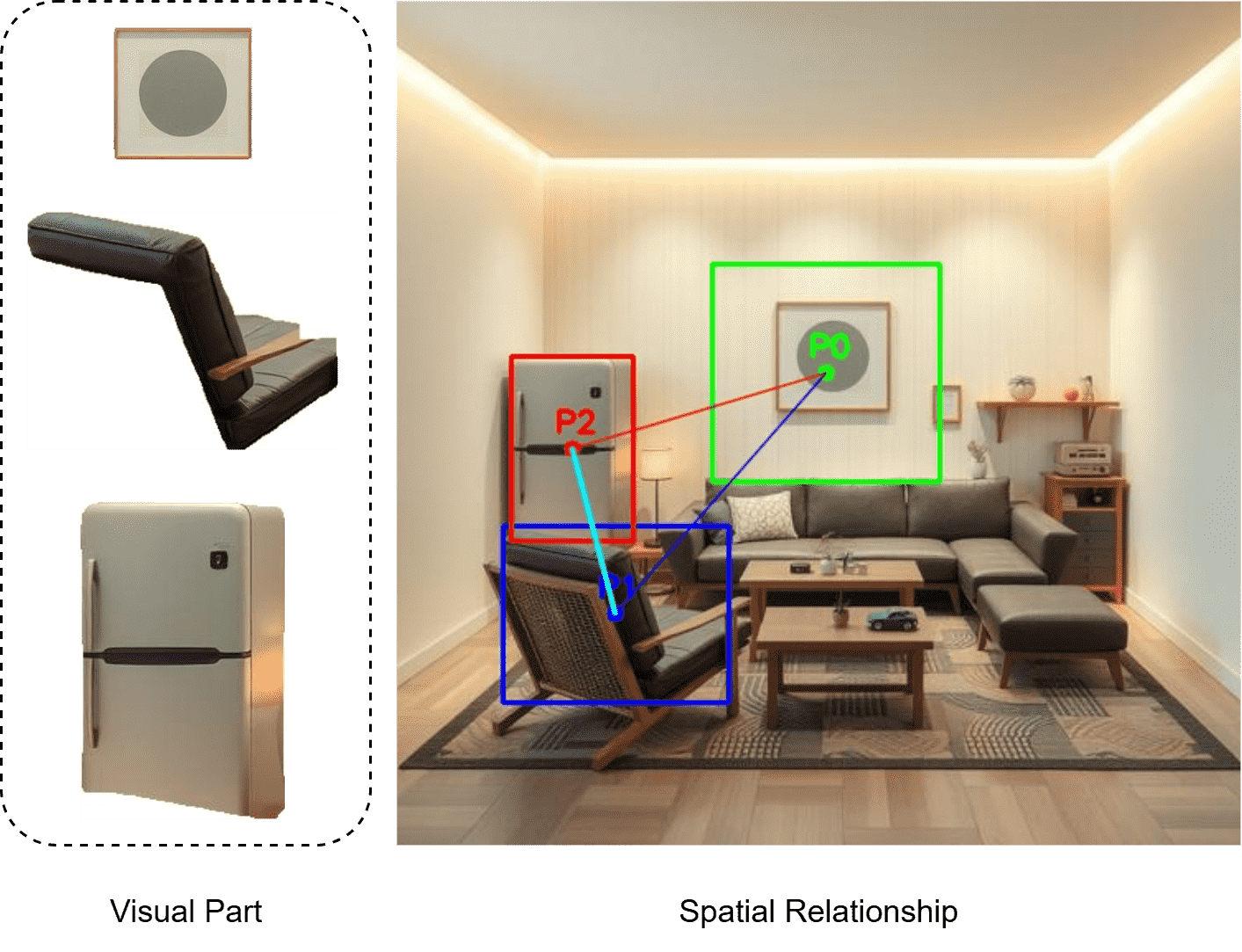}}
\caption{Graph Prior Visualization.}
\label{fig:firstExample}
\end{figure}

\textbf{Motivation for Graph Prior.} PiT treats user-provided parts as an unordered set and therefore ignores the relations that naturally exist between components. We instead augment generation with an explicit graph prior in which each part is a node and edges encode spatial or semantic adjacency, as illustrated in Figure~\ref{fig:firstExample}.

To inject structural awareness, we condition generation on an explicit graph prior $\textbf{G}$. Each part $I_i$ is mapped to a deterministic $IP^+$ embedding $\textbf{h}_i$, and the resulting graph-conditioned distribution factorizes as:
\begin{equation}
\textbf{p}_\theta(\textbf{x} | I_i, \textbf{G}) = \int \textbf{p}_\theta(\textbf{x} | \{\textbf{h}_i\}, A) \prod_{j=1}^N \delta(\textbf{h}_j - IP^+(I_i)) d\textbf{h}_i \label{eq:equation2}
\end{equation}
where the Dirac delta $\delta(\cdot)$ simply expresses that each part embedding $\textbf{h}_i$ is a deterministic output of the pre-trained $IP^+$ encoder rather than a free latent variable; this is the standard encoder–prior factorization. The inner conditional distribution $\textbf{p}_\theta(\textbf{x} \,\vert\, \{\textbf{h}_i\}, \textbf{A})$ is instantiated by our graph-conditioned diffusion prior. The adjacency matrix $\textbf{A} \in \{0,1\}^{N \times N}$ encodes the part relationship graph $\textbf{G}$. Our approach, Graph-PiT, first constructs $\textbf{A}$ from the parts spatial layout (during training) or from user-specified constraints (at inference), then learns a diffusion prior $\textbf{p}_\theta$ conditioned on both $\textbf{h}_i$ and $\textbf{A}$. This mechanism ensures that generated concepts $\textbf{x}$ adhere to the intended part relationships, resolving the unordered-input and coherence challenges inherent in part-based concept composition.

\subsection{Graph Relation Construction from Visual Parts}
During training, we automatically construct a two‑tiered graph $\textbf{G}=(\textbf{V}, \textbf{E}, \textbf{A})$ from each full image and its $N$ segmented visual parts, so that the structural relationships among parts are explicitly encoded. Each visual part $i$ is represented by a super-node $v_i^{super} \in \textbf{V}$, whose feature is its $IP^+$ embedding $Mean(\textbf{h}_i) \in \mathbb{R}^{1 \times D}$. We further expand each super-node into $d$ sub-nodes $v_{i,k}^{sub} \in \textbf{V}$ corresponding to each token of $\textbf{h}_i$, yielding a hierarchical representation. For a more detailed explanation of how to construct the graph nodes and edges of the training data, please refer to Supplementary Section VII.

Edges $\textbf{E}^{super} \in \textbf{E}$ encode spatial or semantic relations between super-nodes (e.g., adjacency, user-provided constraints), while edges $\textbf{E}_i^{sub} \in \textbf{E}$ fully connect each super-node $\textbf{v}_i^{super}$ to its sub-nodes $\textbf{v}_{i,d}^{sub}$. For training data, we implement a graph prior constructor based on bounding box centroids and direction contact encoding. Given the original image and its $N$ visual part crops, our method computes a binary adjacency matrix $\textbf{A}^{super} \in \{0,1\}^{N \times N}$. This super-node adjacency matrix provides an explicit structural prior that informs the subsequent graph-based generative model. At inference time, users instead provide $\textbf{A}^{super}$ directly (for example, via an interactive layout tool), and we reuse the same graph machinery without re-running the alignment or IoU-based construction.

\subsection{Graph-PiT Model Architecture}
\begin{figure}[htbp]
\centerline{\includegraphics[width=.8\linewidth]{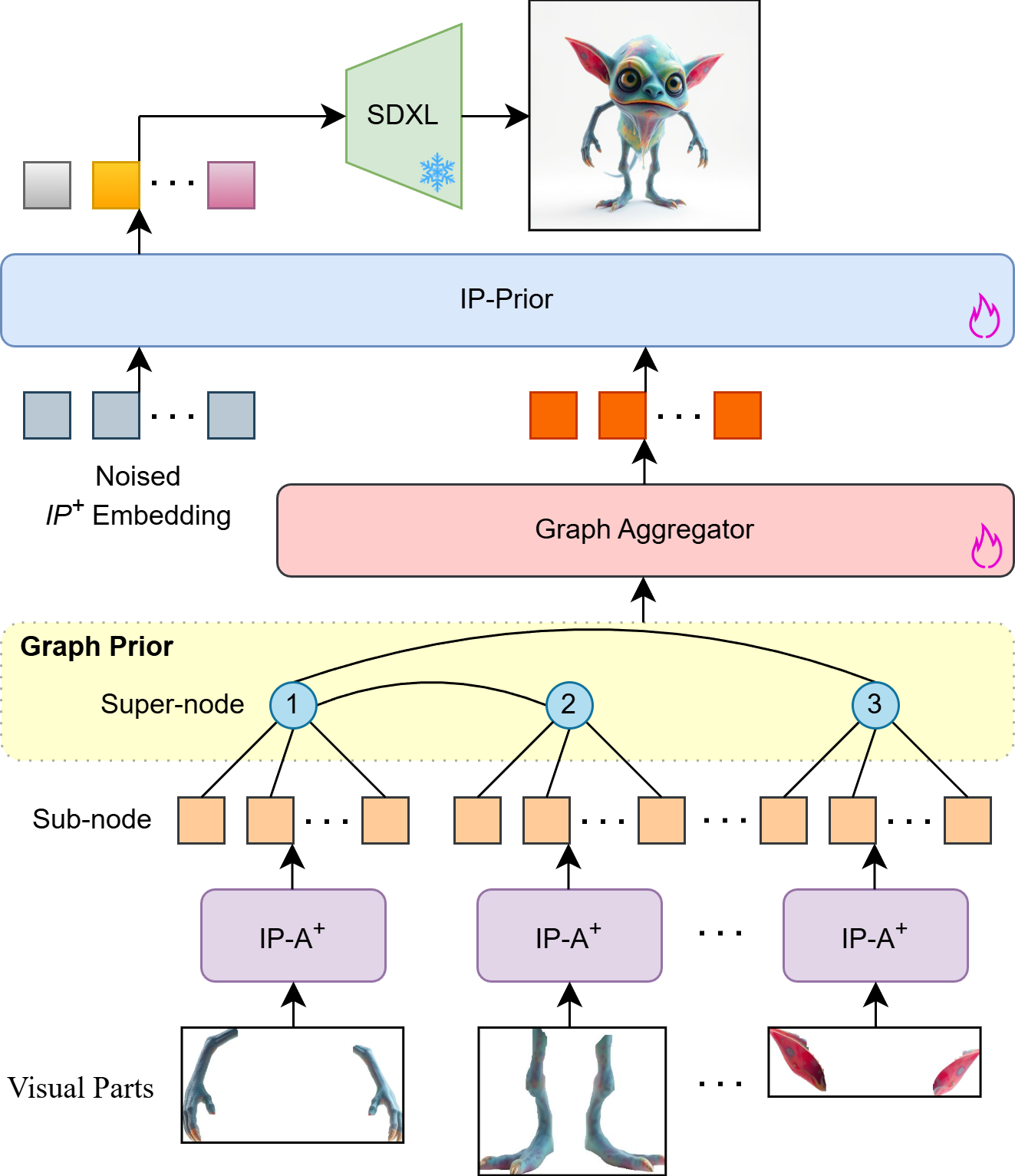}}
\caption{Illustrates the Overall Graph-PiT Pipeline~\cite{richardson2025piece}.}
\label{fig:graph-pit}
\end{figure}

Graph-PiT integrates the graph prior into the PiT IP-Prior through three components: $IP^+$ encoding, a graph aggregator, and a conditional flow-matching prior (Figure \ref{fig:graph-pit}).

A set of part images is first encoded by a pre-trained IP-Adapter$^+$ into $IP^+$ features. These features define the super-nodes, sub-nodes, and adjacency edges of the graph prior. The Graph Aggregator refines them into structure-aware part features, which condition the IP-Prior and are finally decoded by SDXL. At inference time, users may provide any subset of parts together with desired adjacency constraints.

\textbf{IP-Adapter$^+$ Encoding.} Following PiT~\cite{richardson2025piece}, each visual part image $I_i$ is encoded by a pre-trained IP-Adapter$^+$ into the $IP^+$ space:
\begin{equation}
\mathbf{h}_i^{sub} = \mathbf{h}_i = IP\_Adapter^+(I_i) \in \mathbb{R}^{d \times 2048}  \label{eq:equation11}
\end{equation}
where D (e.g., \ 2048) is the embedding dimension. These token features preserve both appearance and semantics and serve as the initial features of the hierarchical graph.
\begin{equation}
\mathbf{X}^{(0)} = \begin{bmatrix}
\mathbf{h}_1^{super} \\
\vdots \\
\mathbf{h}_N^{super} \\
\mathbf{h}_1^{sub} \\
\vdots \\
\mathbf{h}_d^{sub}
\end{bmatrix} = \begin{bmatrix}
Mean(\mathbf{h}_1^{sub}) \\
\vdots \\
Mean(\mathbf{h}_d^{sub}) \\
\mathbf{h}_1^{sub} \\
\vdots \\
\mathbf{h}_d^{sub}
\end{bmatrix} \in \mathbb{R}^{(N+Nd) \times D}  \label{eq:equation12}
\end{equation}

\textbf{Graph Aggregator Module.} The Graph Aggregator is a core component of our model, designed to refine initial part-based feature representations by explicitly modeling the hierarchical and structured relationships between visual parts. Given initial feature embeddings for a set of image patches, the aggregator leverages a predefined graph prior, representing the spatial semantic associations between key components, to guide feature learning. The algorithmic details of the Graph Aggregator are provided in Supplementary Section VIII.

To achieve this goal, we propose a novel Hierarchical Graph Neural Network (HGNN) architecture. This architecture operates at two levels: a high-level super-node graph representing coarse-grained visual components and a low-level sub-node graph representing the fine-grained image patches that comprise these components. This design draws on Graph Convolutional Networks (GCNs) and Graph Attention Networks (GATs), allowing explicit part-level reasoning. This two-layer processing mechanism fosters a rich flow of information: global structural context guides local feature details, while local visual details in turn refine the global semantic representation.

\textbf{Hierarchical Message Passing.} The core of the aggregator is a series of stacked hierarchical graph layer modules. Each layer performs a round of message passing, which consists of three steps: intra-graph aggregation, cross-layer hierarchical attention, and feature update. We use $\mathbf{h}_i^{super(l)}$ and $\mathbf{h}_{ik}^{sub(l)}$ to denote the features of the super-node $v_i^{super}$ and sub-node $v_{ik}^{sub}$ after processing by the $l$-th layer, respectively.

\textit{Intra-Graph Aggregation:} Within each super-node graph, node features are updated independently. We employ a GAT on a super-node graph to capture the varying importance of neighboring super-nodes. The super-node feature update is as follows:
\begin{equation}
\mathbf{h}_i^{super(l)'} = GAT(\mathbf{h}_i^{super(l)}, \{\mathbf{h}_j^{super(l)} | \mathbf{A}_{ij}^{super} = 1\})  \label{eq:equation13}
\end{equation}

We use a standard GCN on the sub-node graph to smooth the features within each semantic component. The sub-node feature update is as follows:
\begin{equation}
\mathbf{h}_{ik}^{sub(l)'} = GCN(\mathbf{h}_{ik}^{sub(l)}, \{\mathbf{h}_{im}^{sub(l)} | (k, m) \in \mathbf{E}^{sub}\})  \label{eq:equation14}
\end{equation}

\textit{Inter-Graph Hierarchical Attention:} This is a key step in achieving bidirectional information flow. To propagate information between super-nodes and their sub-nodes, we use two directed attention mechanisms.

\textit{Top-Down (Super-to-Sub) Update:} Global context is propagated from a super-node to its subordinate sub-nodes to refine local details. For each sub-node $v_{ik}^{sub}$ belonging to super-node $v_i^{super}$, its features are updated based on the features of the super-node $v_i^{super}$. We compute the attention weight
\begin{equation}
\alpha_{ik}^{sub} = \sigma(MLP_{sc}([\mathbf{h}_i^{super(l)'}; \mathbf{h}_{ik}^{sub(l)'}]))  \label{eq:equation15}
\end{equation}
and update
\begin{equation}
\mathbf{h}_{ik}^{sub(l)''} = \mathbf{h}_{ik}^{sub(l)'} + \alpha_{ik}^{sub} \mathbf{h}_i^{super(l)'}  \label{eq:equation16}
\end{equation}
where $[\cdot; \cdot]$ represents the concatenation operation, $\sigma$ is the sigmoid function, and $MLP_{sc}$ is a small multilayer perceptron. The attention weight $\alpha_{ik}^{sub}$ adaptively controls the influence of the super-node context.

\textit{Bottom-to-Top (Sub-to-Super) Update:} Specific features of sub-nodes are aggregated to update the representation of their super-node. The features of a super-node $\mathbf{v}_i^{super}$ are updated by weighted aggregation of the features of all its sub-nodes.
\begin{equation}
\beta_{ik}^{sub} = \sigma(MLP_{cs}([\mathbf{h}_{ik}^{sub(l)'}; \mathbf{h}_i^{super(l)'}])) \label{eq:equation17}
\end{equation}
then
\begin{equation}
\mathbf{h}_i^{super(l)''} = \frac{1}{|d|} \sum_{k \in d} \beta_{ik}^{sub} \mathbf{h}_{ik}^{sub(l)'} \label{eq:equation18}
\end{equation}

This allows the abstract representation of a semantic component to be influenced and modified by the actual visual details of its constituent image patches.

\textit{Feature Update:} Finally, residual LayerNorm updates produce the next-layer features:
\begin{equation}
\mathbf{h}_i^{super(l+1)} = LayerNorm(\mathbf{h}_i^{super(l)''} + \mathbf{h}_i^{super(l)})  \label{eq:equation19}
\end{equation}
\begin{equation}
\mathbf{h}_{ik}^{sub(l+1)} = LayerNorm(\mathbf{h}_{ik}^{sub(l)''} + \mathbf{h}_{ik}^{sub(l)}) \label{eq:equation20}
\end{equation}

\textbf{Conditional Flow-matching Prior.} We condition the prior on the refined sub-node embeddings $\mathbf{H}^{sub}$. Let $\epsilon_\theta$ denote the denoiser; at timestep $t$ with noisy latent $\mathbf{z}_t$,
\begin{equation}
\epsilon_\theta (\mathbf{z}_t, t | \mathbf{H}^{sub}) \label{eq:equation21}
\end{equation}
is implemented via a DIT-style Transformer, which incorporates $\mathbf{H}^{sub}$ through cross-attention blocks:
\begin{equation}
\mathbf{Q} = \text{Embed}(\mathbf{z}_t, t),\quad \mathbf{K} = \mathbf{H}^{sub}\mathbf{W}_K,\quad \mathbf{V} = \mathbf{H}^{sub}\mathbf{W}_V \label{eq:equation22}
\end{equation}
\begin{equation}
\text{Attn}(\mathbf{Q}, \mathbf{K}, \mathbf{V}) = \text{softmax}\left(\frac{\mathbf{Q}\mathbf{K}^T}{\sqrt{d_k}}\right)\mathbf{V} \label{eq:equation23}
\end{equation}

This architecture allows the physical relationship prior to directly attending to structured part representations at every denoising step. When we drop the graph tokens and the associated cross-attention, the module reduces to the original IP-Prior used in PiT~\cite{richardson2025piece}, so Graph-PiT can be seen as a graph-conditioned extension of that prior. Our presented Graph-PiT architecture thus tightly fuses structural graph priors with diffusion-based image synthesis, enabling enhanced compositional control and physically plausible part integration.

\subsection{Loss Function}
In the Graph Aggregator, to ensure that the learned representations are meaningful and adhere to the prior structure of the graph, we introduce two auxiliary loss functions that are jointly optimized with the main generation task.

\textbf{Graph Smoothness Loss.} This loss function is used to enforce the structural prior defined by the super-node graph. It encourages connected supernodes on the graph to have similar feature representations in the latent space. The loss is defined as a Laplace smoothing objective:
\begin{equation}
\mathcal{L}_{\text{smooth}} = \frac{1}{|[i<j]|} \sum_{i<j} \mathbf{A}_{ij}^{\text{super}} \left\| \mathbf{h}_i^{\text{super}(L)} - \mathbf{h}_j^{\text{super}(L)} \right\|_2^2 \label{eq:equation24}
\end{equation}
where $\mathbf{h}_i^{\text{super}(L)}$ is the final super-node feature after $L$-th layers of processing. This loss effectively brings related concepts closer together in the embedding space by minimizing the feature variance along the graph edges.

\textbf{Relational Consistency Loss.} This loss acts as a regularization term to ensure that the final super-node embeddings retain the relational information of the original graph. We train a simple MLP to predict whether an edge exists between two super-nodes based on their concatenated features. For each super-edge $(i,j)$ in the sparsified edge index, we predict a logit:
\begin{equation}
\ell_{ij} = \text{MLP}_{\text{edge}}([\mathbf{h}_i^{\text{super}(L)}; \mathbf{h}_j^{\text{super}(L)}])  \label{eq:equation25}
\end{equation}

The loss function is the binary cross-entropy between the predicted edge probabilities and the true adjacency matrix $\mathbf{A}$:
\begin{equation}
\mathcal{L}_{\text{rel}} = \text{BEC}(\ell_{ij}, \mathbf{A}_{ij}^{\text{super}}) \label{eq:equation26}
\end{equation}

The total auxiliary loss is
\begin{equation}
\mathcal{L}_{\text{graph}} = \lambda_g \mathcal{L}_{\text{smooth}} + \lambda_r \mathcal{L}_{\text{rel}}  \label{eq:equation27}
\end{equation}
which is added to the main generation objective.

\section{Experiments and Results}
\subsection{Experiment Setup}
\textit{Datasets.} We evaluate Graph-PiT on four part-based concept-generation domains: character, product, indoor layout, and jigsaw. Each dataset is generated synthetically using the PiT/Flux-Schnell pipeline and segmented into 1--8 semantic parts. The character, product, indoor-layout, and jigsaw datasets contain 17,038, 14,849, 15,257, and 14,830 images, respectively, with an $80\%/20\%$ train/validation split. Our primary quantitative evaluation follows the PiT synthetic-data protocol; Figure~\ref{fig:figure9} shows qualitative transfer to real web images only.

\textit{Model and training configuration.} Graph-PiT uses two hierarchical graph layers over 2,048-dimensional $IP^+$ tokens. Unless noted otherwise, we use $\lambda_g=1.0$, $\lambda_r=1.0$, $\tau_{IoU}=0.00$, and $\tau_{dist}=512$. Training was performed on a single NVIDIA RTX 4090 (24 GB VRAM) and required roughly eight hours per domain; additional implementation details are provided in Supplementary Section C.

\textit{Evaluation metrics.} We report Fréchet Inception Distance (FID) for image quality and IIS for similarity between the generated image and the reference concept. In ablations, we also report edge-accuracy, i.e., the fraction of input adjacency constraints preserved in the generated image.

\begin{table}[ht]
\centering
\caption{Quantitative comparison.}
\label{tab:table1}
\begin{tabular}{llcc}
\hline
Dataset & Method (year) & FID$\downarrow$ & IIS$\uparrow$ \\
\hline
Character & IP-Adapter$^+$ (2023) & 162.51 & 0.78 \\
                          & $\lambda$-ECLIPSE (2024) & 209.77 & 0.70 \\
                          & OmniGen (2025) & 187.63 & 0.75 \\
                          & PiT (2025) & 191.96 & 0.77 \\
                          & \textbf{Graph-PiT (ours)} & \textbf{95.48} & \textbf{0.88} \\
\hline
Product & IP-Adapter$^+$ (2023) & 119.55 & 0.83 \\
                          & $\lambda$-ECLIPSE (2024) & 121.87 & 0.77 \\
                          & OmniGen (2025) & 142.76 & 0.75 \\
                          & PiT (2025) & 92.87 & 0.79 \\
                          & \textbf{Graph-PiT (ours)} & \textbf{47.90} & \textbf{0.90} \\
\hline
Indoor Layout & IP-Adapter$^+$ (2023) & 319.91 & 0.76 \\
                          & $\lambda$-ECLIPSE (2024) & 318.95 & 0.73 \\
                          & OmniGen (2025) & 382.45 & 0.68 \\
                          & PiT (2025) & 227.70 & 0.81 \\
                          & \textbf{Graph-PiT (ours)} & \textbf{176.72} & \textbf{0.85} \\
\hline                       
Jigsaw & IP-Adapter$^+$ (2023) & 242.29 & 0.58 \\
                          & $\lambda$-ECLIPSE (2024) & 280.23 & 0.61 \\
                          & OmniGen (2025) & 400.93 & 0.62 \\
                          % & PiT (2025) & \textbf{29.75} & \textbf{0.95} \\
                          % & \textbf{Graph-PiT (ours)} & 160.10 & 0.76 \\
                          & PiT (2025) & 206.28 & 0.72 \\
                          & \textbf{Graph-PiT (ours)} & \textbf{160.10} & \textbf{0.76} \\
\hline
\end{tabular}
\end{table}

\begin{table}[ht]
\centering
\caption{Ablation results (reported on a representative character validation subset).}
\label{tab:table2}
\begin{tabular}{lccc}
\hline
Variant & FID$\downarrow$ & Edge-accuracy$\uparrow$ & Notes \\
\hline
Full Graph-PiT & 95.48 & 1.00 & default params \\
w/o Laplacian & 98.09 & 0.98 & $\lambda_g$ = 0 \\
w/o EdgeLoss & 116.01 & 0.80 & $\lambda_r$ = 0 \\
\hline
\end{tabular}
\end{table}

\subsection{Quantitative Comparison}
Table \ref{tab:table1} summarizes the quantitative comparison between Graph-PiT and representative baselines: the original PiT~\cite{richardson2025piece}, IP-Adapter$^+$~\cite{ye2023ip,cao2025ground}, $\lambda$-ECLIPSE~\cite{patel2024lambda}, and OmniGen~\cite{xiao2025omnigen}. We include OmniGen as a representative general-purpose multi-image generator, even though it is not a specialized part-assembly model like PiT. The table lists FID ($\downarrow$) and IIS ($\uparrow$) per dataset.

Across all four domains, Graph-PiT improves both FID and IIS over PiT. Relative to PiT, FID drops from $191.96/92.87/227.70/206.28$ to $95.48/47.90/176.72/160.10$, while IIS rises from $0.77/0.79/0.81/0.72$ to $0.88/0.90/0.85/0.76$. The gains are especially pronounced on the character and product domains, where the part layouts are semantically meaningful and explicit adjacency cues strongly reduce implausible part combinations. On indoor layout, the improvement is smaller but still consistent, suggesting that graph conditioning remains helpful even when the scene structure is more open-ended and spatial variability is higher. Notably, Graph-PiT also surpasses PiT on the challenging jigsaw domain, indicating that the graph prior is beneficial not only for clean semantic parts but also for irregular fragment-like inputs. Graph-PiT also outperforms the other baselines in all four domains. Overall, these results show that explicit graph conditioning improves both perceptual quality and concept fidelity across the synthetic setting.

\subsection{Qualitative Comparison}
\begin{figure*}[htb]
  \centering
  \includegraphics[width=0.8\linewidth]{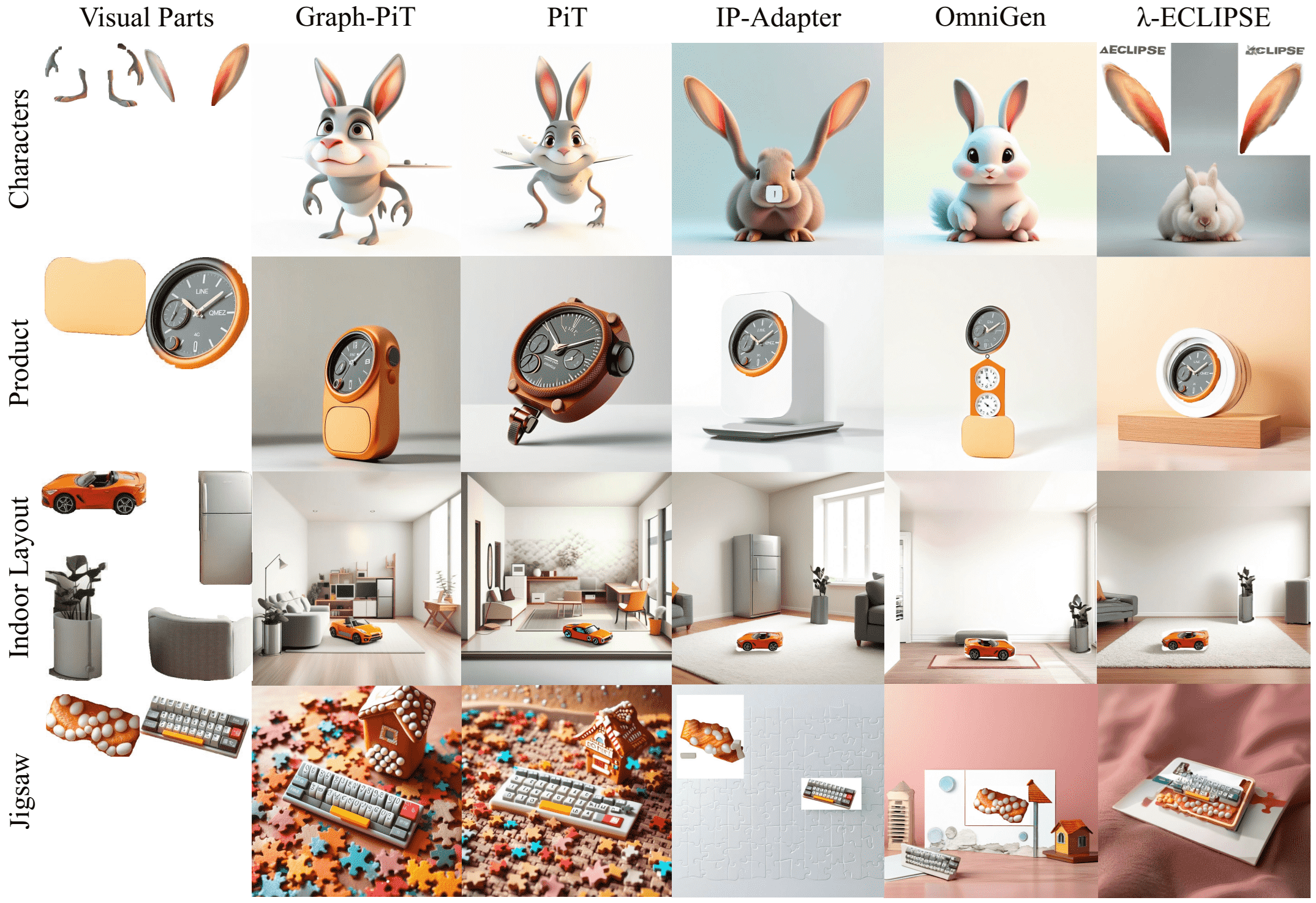}
  \caption{\textit{Qualitative comparison of Graph-PiT results with other models.}}
  \textbf{\label{fig3}}
\end{figure*}

Figure \ref{fig3} presents side-by-side qualitative comparisons on the synthetic benchmark domains. Graph-PiT consistently produces compositions that are more structurally faithful to the input parts and to the provided adjacency constraints. Graph-PiT captures inter-part spatial dependencies that are often overlooked by set-based conditioning approaches, preserving character posture, product assembly, and layout integrity more reliably. PiT and IP-Adapter$^+$ treat parts largely as isolated inputs without explicit pairwise topology, while $\lambda$-ECLIPSE emphasizes latent consistency without structural reasoning. OmniGen serves as a strong general-purpose baseline, but it does not explicitly model part adjacency or topology.

\begin{figure}[htb]
  \centering
  \includegraphics[width=1.0\linewidth]{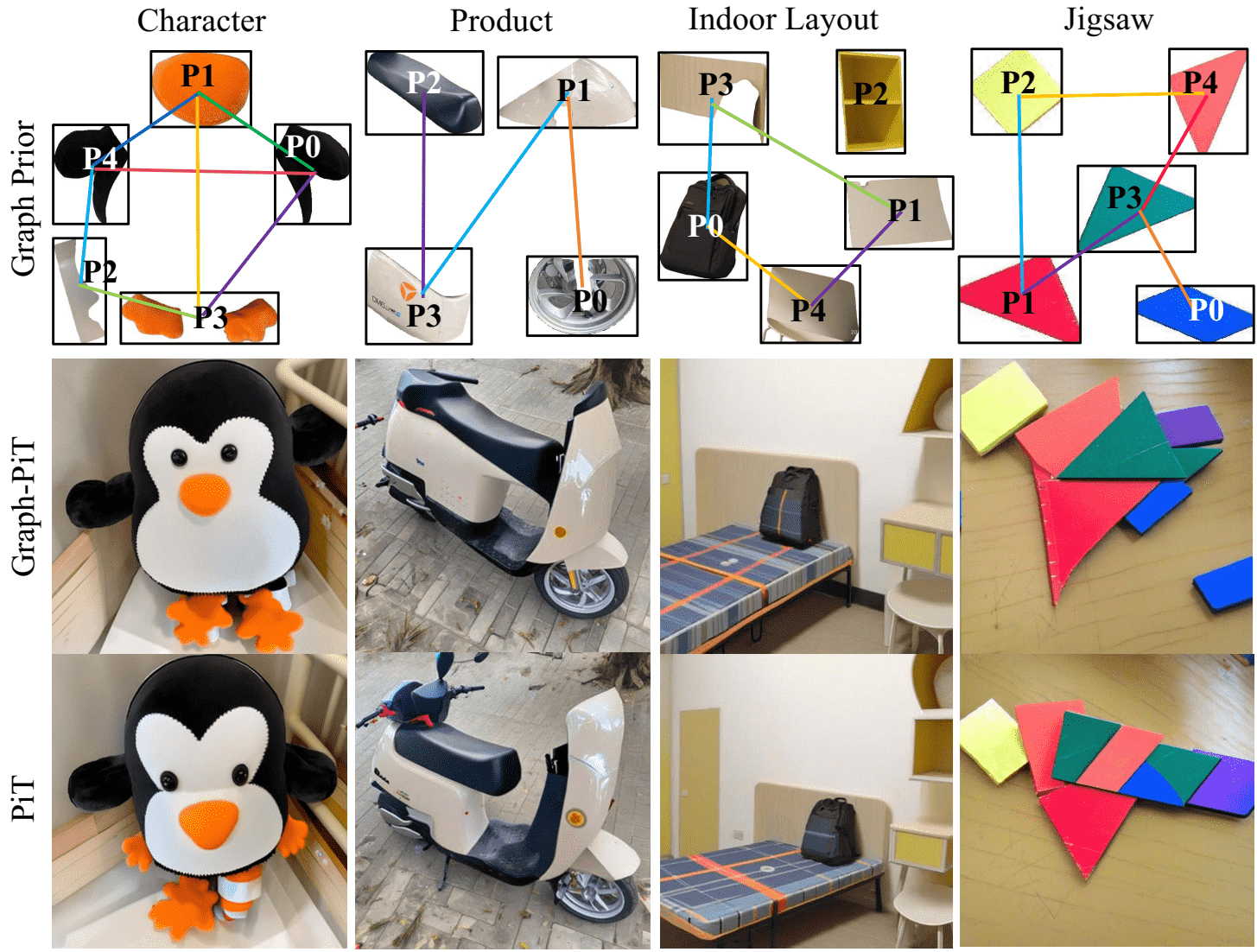}
  \caption{\textit{Qualitative testing of Graph-PiT on real data.}}
  \textbf{\label{fig:figure9}}
\end{figure}

Figure~\ref{fig:figure9} presents qualitative transfer results on real web images after training only on synthetic PiT/Flux-Schnell data. These images are intended as visual checks of transfer behavior rather than quantitative domain-transfer evidence. They reveal a clear synthetic–real gap in photorealism and material detail due to embedding shift, segmentation noise, and decoder mismatch. Despite this, the graph prior continues to place connected parts in physically plausible relative positions, demonstrating strong topological compliance even when appearance degrades. We expect lightweight real-data fine-tuning of IP-Prior/IP-LoRA, stronger augmentations, and improved part localization to reduce this gap while preserving the structural advantages of graph-based part-level control.

\subsection{Ablation Study}
We ablate the two principal structural regularizers in the Graph Aggregator: the Laplacian smoothness loss and the edge-reconstruction loss. Unless otherwise stated, the default setting uses $\lambda_g=1.0$ and $\lambda_r=1.0$, which we selected empirically so that the auxiliary graph objectives remain commensurate with the main generation loss. Table \ref{tab:table2} reports FID and edge-accuracy (fraction of input adjacency constraints satisfied) for each variant. Removing the Laplacian term results in a minor decline (FID 95.48 → 98.09, edge-acc 1.00 → 0.98), suggesting that the Graph Laplacian offers limited smoothing advantages but is not the primary factor influencing structural compliance. On the other hand, removing the edge-reconstruction loss considerably reduces edge-accuracy (1.00 → 0.80) and FID (95.48 → 116.01; $\approx$+21.5\% worse), indicating that explicitly educating the aggregator to maintain adjacency is essential for enforcing component relationships. Additional visualizations are provided in Supplementary Section IX.

\subsection{Limitations}
While Graph-PiT significantly improves compositional fidelity, several limitations remain. Our graph-prior constructor relies on accurate segmentation and bounding-box alignment; heavy occlusion or extremely small parts can yield incorrect adjacency matrices. Graph-PiT currently models only binary adjacency; future work could incorporate richer relation types (e.g., articulation angles or functional semantics). 

\section{Conclusions}
We presented Graph-PiT, a part-based generative model that conditions a diffusion prior on hierarchical graph-aware part embeddings. Across four synthetic domains, including the challenging jigsaw setting, it improves image quality and fidelity to user-provided adjacency constraints. Ablations show that edge reconstruction is a key contributor to these gains. Future work will extend the graph prior to richer relation types and evaluate on real-world part collections.

\section*{Acknowledgment}
This work is supported by Guangdong Provincial Key Laboratory of Ultra High Definition Immersive Media Technology (Grant No. 2024B1212010006).

\bibliographystyle{IEEEtran}
\bibliography{icme2026references}

%-------------------------------------------------------------------------
%Color tables are no longer required for purely electronic publications.
\newpage
\pagebreak[4]
\clearpage

% 在主文档中，参考文献之后
% \appendix
% \setcounter{figure}{0} % 重置图片计数器
% \setcounter{table}{0}  % 重置表格计数器
% \renewcommand{\thefigure}{S\arabic{figure}} % 图片编号格式改为 "S1", "S2"...
% \renewcommand{\thetable}{S\arabic{table}}   % 表格编号格式改为 "S1", "S2"...
% \section{Supplementary Material} % 补充材料的标题
% \textbf{Supplementary Material}
% 这里放入您补充材料.tex文件里的内容
% \input{supplementary.tex} % 使用 \input 引入补充材料文件

\section{Related Works}
\subsection{Controllable Visual Generation}
Modern diffusion models allow rich conditioning via cross-attention~\cite{rombach2022high}. Methods such as DiffusionCLIP~\cite{kim2022diffusionclip} and Attn-and-Excite~\cite{chefer2023attend} use CLIP-based guidance or attention biases to manipulate outputs by text prompts. However, these methods focus on ensuring textual content appears (e.g., exciting attention for each subject~\cite{chefer2023attend}) and do not explicitly model arbitrary input layouts. Similarly, stable diffusion and its extensions (e.g., DreamBooth~\cite{ruiz2023dreambooth}, Kandinsky-style models~\cite{razzhigaev2023kandinsky}) excel at single-object or whole-scene generation but lack built-in mechanisms for assembling user-specified parts with guaranteed consistency. Unlike prompt-based control, we use a structured prior so the model understands which parts must cohere spatially.

\subsection{Part-based and Compositional Models}
There is growing interest in models that are generated by composing meaningful parts. For example, PartGen~\cite{chen2025partgen} addresses 3D generation by first extracting semantically meaningful parts from multi-view inputs using diffusion. PartComposer~\cite{rangwani2025composing} and related work show that artists require fine-grained part-level attributes; existing text-to-image models “either generate an image vastly different or ignore the attribute details”~\cite{rangwani2025composing}. Our work follows this line, but instead of iterative editing or mask-based control, we provide an explicit graph prior. IP-Composer~\cite{dorfman2025ip} and pOps~\cite{richardson2025pops} leverage CLIP embeddings to blend concepts~\cite{richardson2025piece}, but do not account for spatial relations between parts. Graph-PiT extends this by embedding part images in a latent space (e.g., via an IP-Adapter~\cite{richardson2025piece}) and injecting relational constraints via a graph GNN.

\subsection{Layout and Scene Graph Generation}
Scene graph to image models~\cite{johnson2018image} encode object relations into images: Johnson et al. proposed using graph convolutions on a scene graph to predict object layouts and refine images accordingly~\cite{johnson2018image}. More recent 3D layout work (e.g., DiffuScene~\cite{tang2024diffuscene}) diffuses unordered sets of object attributes to synthesize room layouts. These methods operate on abstract object categories and bounding boxes; in contrast, Graph-PiT works on user-provided image fragments (parts) and their explicit graph. Layout-guided T2I models (e.g., InstructScene~\cite{lin2024instructscene}) also use graph diffusion for 3D scene generation, highlighting the benefit of graph priors. Our approach bridges these ideas: we apply graph neural modeling directly to the user’s part embeddings, enabling image generation that honors the specified structure.

\section{Preliminaries: $IP^+$ Representation and GNN Background}
\subsection{$IP^+$ Embedding and PiT Overview}
Following Richardson et al.~\cite{richardson2025piece}, PiT operates in an IP-Adapter$^+$ representation space. Given a visual part image $I_i$, a pre-trained IP-Adapter$^+$ encoder produces a token grid $\mathbf{H}_i \in \mathbb{R}^{d \times D}$ (for example, $d{=}16$ tokens of dimension $D{=}2048$), where each token encodes local appearance and semantics. The original PiT pipeline then concatenates the part-token sets and feeds them to an IP-Prior generative model, which synthesizes a joint embedding that is decoded into the final image. Graph-PiT preserves this PiT pipeline but inserts a graph-conditioned aggregator between the encoder and the prior, so that part relationships can be modeled explicitly rather than treating the parts as an unordered set.

Concretely, for each part image $I_i$, we obtain an $IP^+$ token grid $\mathbf{H}_i$ whose rows correspond to spatially arranged visual descriptors. These tokens are later interpreted as sub-nodes in our hierarchical graph, while their mean feature gives a part-level ``super-node'' representation. This view is compatible with the original PiT design but prepares the ground for injecting structural information via graph message passing.

\subsection{Graph Neural Networks in Brief}
Graph Neural Networks (GNNs) update node features by aggregating information from their neighbors along edges in a given graph, making them well-suited to encode relational inductive biases. Early works like Kipf and Welling’s GCN~\cite{kipf2016semi} learn node embeddings that integrate local graph structure. Recent latent diffusion on graphs~\cite{zhou2024latent} shows that one can simultaneously generate graph topology and features in a continuous space. We adopt a hierarchical GNN (nodes for parts and sub-part features) that propagates information along edges. Auxiliary losses (e.g., a Laplacian regularizer~\cite{pang2017graph}) encourage smoothness over the graph and structural consistency. In this way, Graph-PiT leverages relational inductive biases via GNNs. It fuses graph-based reasoning (as in scene graphs~\cite{johnson2018image}) with the power of modern diffusion priors~\cite{rombach2022high}.

In Graph-PiT, nodes correspond to parts (and their $IP^+$ tokens) while edges encode spatial or semantic adjacency. These basic ingredients are organized into a hierarchical design over super-nodes and sub-nodes tailored specifically to part-based image generation. Super-nodes exchange information over the part-level graph (capturing which components should be adjacent), while sub-nodes propagate fine-grained appearance information within each part. Bidirectional top-down and bottom-up message passing then couples these two levels, allowing global structural context to guide local token features and vice versa. We rely on standard message-passing graph neural networks (GCN/GAT-style layers)~\cite{kipf2016semi} to encode relational inductive biases over the part graph, which perform neighborhood aggregation followed by a learned linear transformation and nonlinearity.

The main paper focuses on this hierarchical architecture and its role in enforcing part relationships; readers seeking more background on generic GNN formulations can refer to standard surveys and tutorials.

\section{Training Data and Generation}
\subsection{Graph Relation Construction from Visual Parts}
\textbf{Node Representation.} In Graph-PiT, visual parts are encoded at two levels of granularity, super-nodes and sub-nodes, to capture both holistic and fine-grained information from the $IP^+$ embedding space.

Given $N$ input part images $\{I_1, I_2, \ldots, I_N\}$, we encode each visual part $I_i$ into a $d \times D$-dimensional dense feature token $\textbf{h}_i$ using the IP-Adapter$^+$ encoder.
\begin{equation}
\mathbf{h}_i = IP\_Adapter^+(I_i) \in \mathbb{R}^{d \times D} \label{eq:equation3}
\end{equation}

To enable fine-grained interaction among parts, we decompose $\mathbf{h}_i$ into its individual dimensions, creating $d$ sub-nodes per part $\textbf{v}_{i,k}^{sub} \in \mathbf{V}, k = 1, \ldots, d$ with each sub-node carrying a one-dimensional token $\mathbf{h}_{i,k}^{sub} = [\mathbf{h}_i]_k \in \mathbb{R}^{1 \times D}$. Collectively, the sub-node set for part $i$ is $\{\mathbf{v}_{i,1}^{sub}, \mathbf{v}_{i,2}^{sub}, \ldots, \mathbf{v}_{i,d}^{sub}\}$.

Most importantly, we regard each visual part $I_i$ as a super-node $\mathbf{v}_i^{super} \in \mathbf{V}$. The link relationship between them is the graph prior proposed in this paper, which we will elaborate on in the next subsection. We link each super-node $\mathbf{v}_i^{super}$ to its sub-nodes via intra-part edges, forming a star topology.
\begin{equation}
\mathbf{E}^{sub} = \{(\mathbf{v}_i^{super}, \mathbf{v}_{i,k}^{sub}) | i = 1 \ldots N, k = 1 \ldots d\} \label{eq:equation4}
\end{equation}
\begin{equation}
\mathbf{A}^{sub} = \begin{cases}
\mathbf{A}_{0j} = \mathbf{A}_{j0} = 1 (j = 1, \ldots, d) \\
\mathbf{A}_{ij} = 0 (i, j \geq 1) \\
\mathbf{A}_{ii} = 0
\end{cases}  \label{eq:equation5}
\end{equation}

The attribute of a super-node is the average of the attributes of all its linked child nodes.
\begin{equation}
\mathbf{h}_i^{super} = Mean(\mathbf{h}_i) = \frac{1}{d}\sum_{k=1}^d \mathbf{h}_{i,k} \in \mathbb{R}^{1 \times D} \label{eq:equation6}
\end{equation}

This hierarchical arrangement allows the model's graph layers to propagate information both between parts (via super-nodes) and within each part's embedding (via sub-nodes). By combining the global visual part super-node $V^{super}$ with local feature dimensions $V^{sub}$, our node representation supports downstream graph message passing that jointly reasons over inter-part relationships and intra-part feature interactions. This rich hierarchical encoding is critical for capturing the structural nuances required for coherent part-based concept generation.

\textbf{Edge Definition.} Having defined the nodes, we now specify how the edges $\mathbf{E}^{super}$ and link matrix $\mathbf{A}^{super}$ of the super-nodes are constructed to form the graph prior. The specific process includes component localization, contact determination, orientation encoding, and link matrix generation. Each edge encodes whether two visual parts should be considered adjacent, that is, "touching" or "overlapping", based on their recovered positions in the original image. We proceed in three steps: (1) recovering the coordinates of the visual parts via affine alignment, (2) extracting geometric features (bounding boxes and centroids), and (3) constructing a binary adjacency matrix using IoU and centroid distance thresholds.

\textbf{Affine Alignment and Part Localization.} Each part crop may be arbitrarily scaled and rotated relative to the full image. We estimate an affine transformation $\mathbf{T}_i: (x, y) \to (x', y')$ using feature matching (e.g., SIFT and RANSAC) or fallback template matching. Applying $T_i$ to the four corners of the part image yields transformed corner coordinates $\{x'_{i,n}, y'_{i,n}\}_{n=1}^4$ in the original-image coordinate frame. From these, we compute the part's axis-aligned bounding box
\begin{equation}
b_i = (x_i^{min}, y_i^{min}, w_i, h_i)  \label{eq:equation7}
\end{equation}
 
The centroid of the part is then
\begin{equation}
c_i = (c_i^x, c_i^y) = \left(\frac{x_i^{min} + x_i^{max}}{2}, \frac{y_i^{min} + y_i^{max}}{2}\right)  \label{eq:equation8}
\end{equation}
where $x_i^{min} = \min(x'_{i,n})$, $x_i^{max} = \max(x'_{i,n})$, $w_i = x_i^{max} - x_i^{min}$ and similarly, for $y_i^{min}, y_i^{max}, h_i$.

\textbf{Geometric Feature Extraction.} Using the localized bounding boxes and centroids, we define two pairwise measures for parts $i$ and $j$: Intersection over Union (IoU) and Centroid Euclidean Distance.
\begin{equation}
IoU(b_i, b_j) = \frac{area(b_i \cap b_j)}{area(b_i) + area(b_j) - area(b_i \cap b_j)}  \label{eq:equation9}
\end{equation}

\textbf{Adjacency Matrix Construction.} We fix two thresholds, $\tau_{IoU}$ and $\tau_{dist}$. The visual parts $I_i$ and $I_j$ are declared adjacent if they either overlap sufficiently or lie sufficiently close:
\begin{equation}
\mathbf{A}_{ij}^{super} = \begin{cases}
1, & i \neq j \text{ and } (IoU(b_i, b_j) \geq \tau_{IoU}) \vee (d_{ij} \leq \tau_{dist}) \\
0, & \text{otherwise}.
\end{cases}
\label{eq:equation10}
\end{equation}

The diagonal elements $\mathbf{A}_{ii}$ are always 0. This adjacency matrix $\mathbf{A}^{super}$ serves as the graph prior for the super-node layer in Graph-PiT. By explicitly encoding 2D spatial contacts and overlaps, we constrain the generative model to respect visible part adjacency and topology, which is the scope of structural coherence considered in this paper.

\subsection{Training data generation: Prompt} 
The different scene data used in this paper follow PiT, generated using Flux-Schnell and specific prompts. We have added the interior layout and jigsaw scenes, while the character and product design prompts are still consistent with PiT. The following are the prompts used for them:
\begin{itemize}
    \item Indoor Layout Prompt: \textit{"An indoor room layout design photo showing a \{attrib-utes\_and\_materials\_txt\} room with \{character\_txt\} furniture attributes (sofa, chair, refrigerator, table, toy car, etc.) neatly arranged and perfectly integrated to form a comfortable and cozy Japanese home scene. The photo is set against a brightly lit background with soft gradient colors, creating a neutral and elegant atmosphere that highlights the minimalist design of the modern home. The soft and even lighting highlights the outlines and textures, giving the composition a professional, sophisticated quality."}
    \item Jigsaw Prompt:  \textit{"A jigsaw picture showing a diverse \{attributes\_and\_materials\_txt\} jigsaw game scene, with \{character\_txt\} brightly colored building blocks or pieces or product parts scattered around yet forming a whole."}
\end{itemize}

\begin{figure}[htb]
  \centering
  \includegraphics[width=\linewidth]{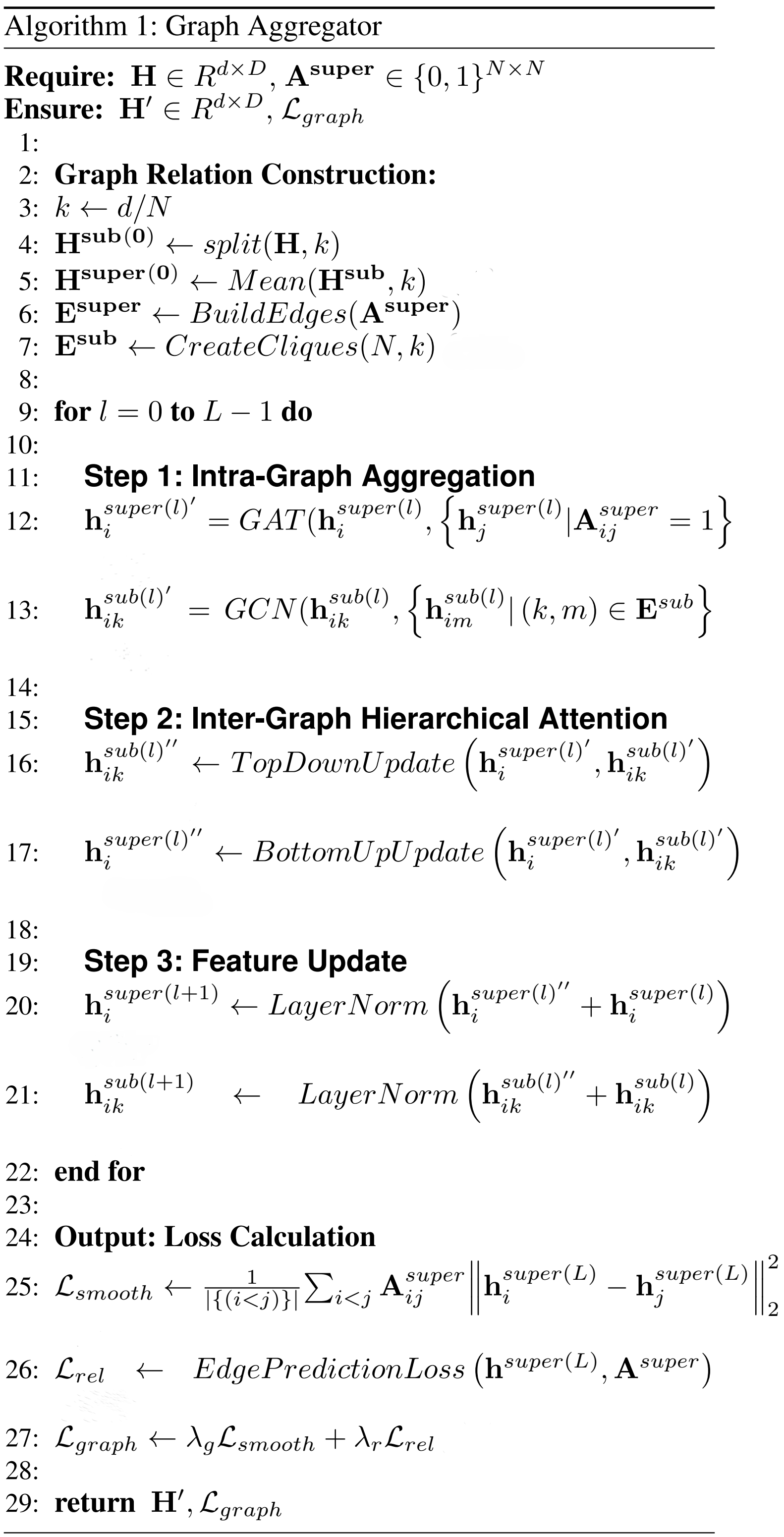}
\end{figure}

\subsection{Reproducibility and Implementation Summary} 
To facilitate replication, our quantitative protocol follows the same synthetic-data setup as PiT for the character, product, indoor-layout, and jigsaw domains. For the camera-ready version, we re-ran the PiT baseline on jigsaw with the corrected evaluation pipeline and obtained FID 206.28 and IIS 0.72; Graph-PiT remains better on this domain with FID 160.10 and IIS 0.76. Unless otherwise stated, the reported Graph-PiT setting uses two hierarchical graph layers on top of the PiT pipeline, with graph-construction thresholds $\tau_{IoU}=0.00$ and $\tau_{dist}=512$ pixels and default auxiliary-loss weights $\lambda_g=1.0$ and $\lambda_r=1.0$. In our training configuration, images are generated at 1024$\times$1024 resolution, with batch size 4, gradient accumulation 4, learning rate $1\times10^{-5}$, mixed-precision FP16, and 10k training steps on a single NVIDIA RTX 4090 (24 GB VRAM). We include these values here as a compact implementation summary so that readers can more easily reproduce the camera-ready results.

\section{Algorithm Details}
The pseudo-code (Algorithm 1) summarizes our Graph Aggregator’s forward pass. In prose, each major stage corresponds to
\begin{itemize}
    \item We first split the flat token sequence $\mathbf{H} \in \mathbb{R}^{d \times D}$  into $N$ groups of $k = d/N$ sub-node embedding (Line 4), then average each group to form the initial $N$ super-node embedding (Line 5).
    \item We convert the binary super-node adjacency $\mathbf{A}^{super}$ into an edge list $\mathbf{E}^{super}$ (Line 6) and build the corresponding intra-part super-to-sub edges $\mathbf{E}^{sub}$ (Line 7).
\end{itemize}

For each of the $L$ graph layers, we perform three steps:
\begin{itemize}
    \item Super-node update (Line 12): we run a graph-attention (GAT) convolution over $\mathbf{E}^{super}$ to produce updated super-node features $\mathbf{h}_i^{super(l)'}$.
    \item Sub-node update (Line 13): we run a standard GCN convolution over the sub-node intra-part edges $\mathbf{E}^{sub}$, yielding $\mathbf{h}_{ik}^{sub(l)'}$.
    \item Top-down update (Line 16): each sub-node attends to its super-node via a small MLP, injecting high-level context into $\mathbf{h}_{ik}^{sub(l)''}$.   % (formally Equations \ref{eq:equation15}-\ref{eq:equation16}) \eqref{eq:equation16}
    \item Bottom-up update (Line 17): each super-node aggregates signals from its $k$ sub-nodes to form $\mathbf{h}_i^{super(l)''}$. % (Equations \ref{eq:equation17}-\ref{eq:equation18})
    \item To obtain the inputs for the next layer, we implement residual connections and then layer normalization to both super- and sub-nodes (Lines 20-21).
    \item After $L$ layers, we calculate a Laplacian smoothness term $\mathcal{L}_{\text{smooth}}$ (Line 25 / Equation \ref{eq:equation24}) by averaging squared differences between neighboring super-nodes. Next, we implement a lightweight MLP to predict each super-edge and calculate a relation-reconstruction cross-entropy loss $\mathcal{L}_{\text{rel}}$ (Line 26). %/ Equation \ref{eq:equation26}
    \item We then define $\mathbf{H}' = \mathbf{H}^{sub(L)}$, i.e., we take the refined sub-node embeddings as the graph-conditioned token set passed to the conditional prior.
    \item Finally, we combine these as $\mathcal{L}_{\text{graph}} = \lambda_g \mathcal{L}_{\text{smooth}} + \lambda_r \mathcal{L}_{\text{rel}}$ (Line 27). %/ Equation \ref{eq:equation27}
    \item The module returns the refined sub-node embedding $\mathbf{H}^{\prime}$ and the accumulated structural loss $\mathcal{L}_{\text{graph}}$.
\end{itemize}

This algorithm ensures that the Graph Aggregator both improves part-level feature coherence according to the user-provided graph prior and supervises the model to respect the intended adjacency structure.

\begin{figure*}[htb]
  \centering
  \includegraphics[width=1.0\linewidth]{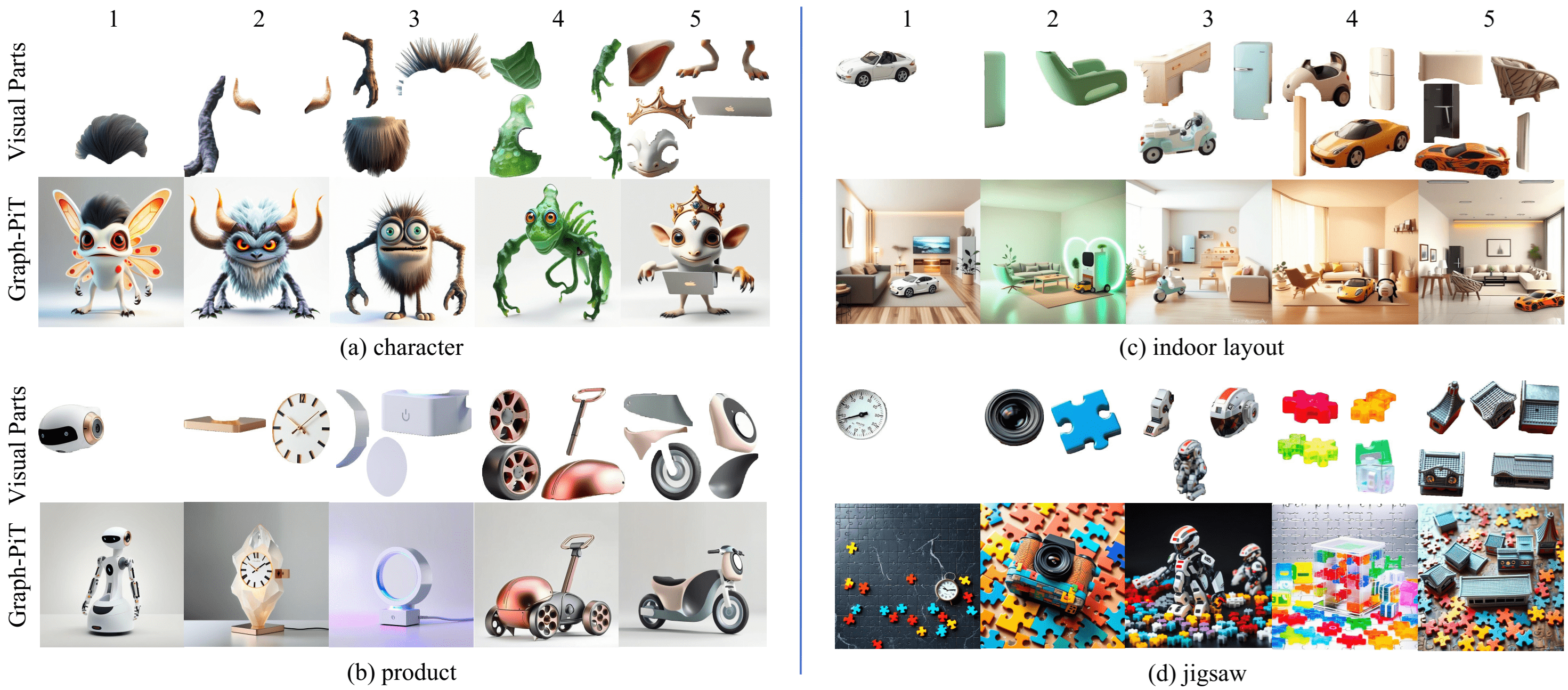}
  \caption{\textit{Qualitative comparisons were made of the generated results of Graph-PiT with different numbers of visual components input on character, product, interior layout, and puzzle datasets.}}
  \textbf{\label{fig5}}
\end{figure*}

\section{Qualitative Ablation Study} 
Beyond the quantitative ablations in the main paper, we provide qualitative examples across four domains---character, product, indoor layout, and jigsaw---to illustrate how Graph-PiT behaves as the number of conditioned parts increases from one to five. Figure \ref{fig5} should be read primarily as a structural visualization: it shows whether the generated image respects the intended part adjacency and overall topology when increasingly informative graph priors are supplied.
\begin{itemize}
    \item Figure \ref{fig5} (a) (character): with only one part, Graph-PiT reduces to a weakly constrained PiT-like setting and may invent extra anatomy around the provided fragment. As more parts are added (e.g., head, torso, wings), the graph prior stabilizes the global topology and places the components in more semantically consistent positions.
    \item Figure \ref{fig5} (b) (product): when users provide fragments such as wheels, housings, or handles, Graph-PiT better preserves relative orientation and attachment structure, especially once several complementary parts are available.
    \item Figure \ref{fig5} (c) (indoor layout): for furniture placement, the graph prior helps maintain plausible local adjacency (e.g., a chair near a table or a sofa against a wall) while the diffusion prior fills in the broader room context.
    \item Figure \ref{fig5} (d) (jigsaw): this domain remains the most challenging because the pieces are irregular and often only weakly informative in isolation. Even so, the corrected quantitative evaluation still favors Graph-PiT over PiT, and the qualitative examples show that graph conditioning provides useful topological cues under noisy part decompositions.
\end{itemize}

Across the four controlled synthetic domains used in our quantitative evaluation---character, product, indoor layout, and jigsaw---stronger graph supervision generally improves adjacency preservation and reduces implausible placements. The jigsaw examples further show that hierarchical graph conditioning remains helpful even under substantially noisier part decompositions.

% \bibliographystyle{IEEEbib}
% \bibliography{icme2026references}
\end{document}